\documentclass[english]{article}
\usepackage[T1]{fontenc}
\usepackage[latin9]{inputenc}
\usepackage{mathtools}
\usepackage{amsmath}
\usepackage{amssymb}
\usepackage{stackrel}
\usepackage{esint}
\usepackage{nameref}

\makeatletter
\numberwithin{equation}{section}

\usepackage{arxiv}
\usepackage[T1]{fontenc}    % use 8-bit T1 fonts
\usepackage{hyperref}       % hyperlinks
\usepackage{url}            % simple URL typesetting
\usepackage{booktabs}       % professional-quality tables
\usepackage{amsfonts}       % blackboard math symbols
\usepackage{nicefrac}       % compact symbols for 1/2, etc.
\usepackage{microtype}      % microtypography
\usepackage{graphicx}

\makeatother

\usepackage{babel}
\begin{document}
\title{Surface Segmentation Using Implicit Divergence Constraint Between
Adjacent Minimal Paths}
\author{Jozsef Molnar\\Synthetic and Systems Biology Unit 
\\Biological Research Centre (BRC)
\\Szeged, Hungary
\\
\texttt{jmolnar64@digikabel.hu} \And
Peter Horvath
\\Synthetic and Systems Biology Unit
\\Biological Research Centre (BRC) 
\\Szeged, Hungary
\\
\\Institute for Molecular Medicine Finland-FIMM
\\Helsinki Institute of Life Science-HiLIFE 
\\University of Helsinki
\\Helsinki, Finland
\\
\\Single-Cell Technologies Ltd
\\Szeged, Hungary
\\
\texttt{horvath.peter@brc.hu} \hypersetup{ pdftitle={Surface Segmentation Using Implicit Distance Constraint Between Adjacent Minimal Paths}, pdfauthor={Molnar, J.}, pdfkeywords={3D object extraction, minimal surface Eikonal equations, divergence constraint }, } }
\date{November 11. 2021}
\maketitle
\begin{abstract}
We introduce a novel approach for object segmentation from 3D images
using modified minimal path Eikonal equation. The proposed method
utilizes an implicit constraint - a second order correction to the
inhomogeneous minimal path Eikonal - preventing the adjacent minimal
path trajectories to diverge uncontrollably. The proposed modification
greatly reduces the surface area uncovered by minimal paths allowing
the use of the calculated minimal path set as ``parameter lines''
of an approximate surface. It also has a loose connection with the
true minimal surface Eikonal equations that are also deduced.

\end{abstract}

\section{Introduction}

The active contour models have proven record of success in various
image segmentation problems. Since its inception \cite{kass1988snakes}
the method evolved in many directions. From theoretical point of view,
the most prominent changes were: a) the switch from the parametric
models to the geometric \cite{Yezzi,caselles1997geodesic,caselles1997minimal},
where the functional energy does not depend on the parameterization,
only on the image content and the intrinsic geometric properties of
the contour and b) the transition from the Lagrangian approach to
the Hamilton-Jacobi. In the former case, the solution is obtained
via the Euler-Lagrange equation associated with the functional, whereas
in the latter, the Eikonal (the ``time independent'' Hamilton-Jacobi)
equation needs to be solved for the minimal action map \cite{Cohen97globalminimum,ShortestpathKimmel}.\footnote{In the minimal path framework, the minimal action map can be identified
with distance map wrt a distance function induced by metric. For this
reason, in this paper we use the \textit{minimal action map} and the
\textit{distance map} interchangeably. } The Hamilton-Jacobi approach has also turned the view from the ad
hoc interpretation of the segmentation functionals to the geodesic
calculation in a Riemannian manifold wrt the metrics the functional
represent, placing the segmentation problems into one consistent theoretical
framework. Beyond the theoretical aspects, the biggest advantage of
the Hamilton-Jacobi formulation compared to the Lagrangian - is that
it always provides unique, globally minimal segmentation avoiding
local minima traps, the Lagrangian approach often suffers from; and
this is without significant - if any - increase in run-time. It is
also worth to mention that despite the fact that the Eikonal equations
can only be derived from functionals having only the first order quantity
of the path geometry - for the price of additional dimension(s) -
higher order quantities, like the Euler's elastica \cite{mumford1994elastica}
can also be modeled \cite{DBLP:journals/ijcv/ChenMC17} as well as
problems, formulated with region energy \cite{chan2001active}-\cite{chen2016finsler,chen2019eikonal}.
An extensive summary of the geodesic methods can be found in \cite{peyre:hal-00528999}.

However, despite the great success in the field of standard 2D image
segmentation, the Hamilton-Jacobi approach cannot be directly transferred
into three dimensional problems due to the essentially different nature
of the minimal surface equations. The imparity is the consequence
of the fact that the minimal surface is determined by the two dimensional
sub-determinant, whilst the minimal path functionals rely only on
the one-dimensional minor of the space-metric tensor - an obvious
loss in the information that would be needed for a theoretically correct
3D object segmentation with minimal surfaces. This information hiatus
is manifested in the discontinuity in the minimal path network \cite{articleMantegazza}.
Attempts have been carried out to mitigate the - often devastating
- effect of the metric-information loss, one of the most promising
one being the exploitation of the transport equations \cite{ardon2006fast}.
In this paper we follow a different approach that attempts to expand
the Eikonal equation with an implicit constraint preventing the development
of high divergence bulges on the momentary distance map boundary during
its construction. This in turn penalizes the distances growing uncontrollably
high between the adjacent minimal paths. It will also be shown, that
the constraint we used, has loose connection with the minimal surface
Eikonal equation derived in the \nameref{sec:Appendix}.

\section{Segmentation with Minimal Paths}

Segmenting images in the variational framework is the task of finding
extremal curves of a specifically designed functional that takes its
extrema at the object boundaries corresponding to a logical partitioning
of the image. In the image processing, two major approaches are used
to find these extremals: a) solving the associated Euler-Lagrange
ODE system or b) using the Hamilton-Jacobi PDEs. Both have advantages
and drawbacks. Euler-Lagrange equations can relatively be easy to
derive for functionals incorporating higher order derivatives of the
sought contours or for multiple integrals \textit{i.e.} for boundary
surface determination in 3D voxel images. On the other hand, using
the Euler-Lagrange equations requires initial contour (2D) or surface
(3D) definition close to the solution otherwise the evolving contour
can be trapped in local minima. Defining the whole initial contour,
let alone a surface is a slow, tedious task. 

In contrast, wherever applicable the use of the Hamilton-Jacobi equations
is preferred. It always provides globally optimal solution that is
the shortest distance \textit{minimal path} between any two points
wrt the enforced metric (see Appendix A1-A2). Minimal user input,
a few - often only two - points in 2D or few ``zero set'' contours
in 3D are sufficient. On the other hand, the direct use of the Hamilton-Jacobi
equation is limited to functionals of one-variable with Lagrangian
composed from the zeroth and first - as the highest - order derivatives.
However, incorporating higher-order derivatives is not impossible
for the price of additional dimension(s), more complex Lagrangian
and approximate nature. An efficient such extension is found in \cite{DBLP:journals/ijcv/ChenMC17}
that uses Euler's elastica modeling ``turning inertia''. The method
can be used to segment tubular structures (blood vessels) from X-ray
images correctly handling crossings, bifurcations. Unfortunately,
the extension of the Hamilton-Jacobi theory to the double integral
\textit{minimal surface} problem leads to a completely different problem:
minimal action (\textit{i.e.} distance) map creation in the space
of the function-triplets (see Appendix \nameref{subsec:A.3-Minimal-Surface}).
In this paper we do not pursue the solution in this infinite dimensional
space. Rather we propose an extension of the classical minimal path
Eikonal equation with an implicit constraint that has some (albeit
loose) connection with the true minimal surface problem. 

In three dimensions, functionals, realizing \textit{isotropic inhomogeneous}
metric 
\begin{gather}
\int\phi\left(\mathbf{r}\right)\left|\dot{\mathbf{r}}\right|dt,\:\mathbf{r}\left(t\right)\in\mathbb{R}^{3}
\end{gather}
are almost exclusively used. This metric is the special form of the
Riemannian metrics $\int\sqrt{\left(\dot{\mathbf{r}}\cdot\mathbf{G}\cdot\dot{\mathbf{r}}\right)}\,dt$
where the metric tensor is proportional to the identity tensor $\mathbf{G}=\phi^{2}\left(\mathbf{r}\right)\mathbf{I}$.
The $\phi$ is a carefully defined function usually of the image intensity
gradient $\nabla I$, representing edges. The typical choices are
$\frac{1}{\varepsilon+\left|\nabla I\right|^{p}}$, $\exp\left(-\beta\left|\nabla I\right|\right)$
that take small values at high gradient magnitudes. Isotropic metrics
have the useful property that their minimal paths (the geodesics wrt
the metric) are always perpendicular to the level sets of the distance
function the metric induce. The proposed divergence constraint relies
on this property.

\section{The Minimal Path Network and the Minimal Surface Problem}

\label{sec:The-Minimal-Path-net}

The inspiration of this paper is as follows. 

To emphasize the connection with the minimal surface problem, in this
section we use the following notations: let $\mathbf{s}_{1},\,\mathbf{s}_{2}$
be two closed curves (hereinafter \textit{contours}) of $\mathbb{R}^{3}$.
The path network $\mathcal{N}_{\mathbf{s}_{1}}^{\mathbf{s}_{2}}$
introduced in \cite{ardon2006fastconstrained} contains paths $\mathbf{S}_{\mathbf{s}_{1}}^{q}\left(u\right)$
between the points of $\mathbf{s}_{2}$ and $\mathbf{s}_{1}$:
\begin{equation}
\mathcal{N}_{\mathbf{s}_{1}}^{\mathbf{s}_{2}}=\left\{ \mathbf{S}_{\mathbf{s}_{1}}^{q}\right\} _{q\in\mathbf{s}_{2}}.
\end{equation}
An individual path's Riemannian length wrt an isotropic inhomogeneous
metric is defined by the functional $\int\phi\left(\mathbf{S}_{\mathbf{s}_{1}}^{q}\right)\left|\dot{\mathbf{S}}_{\mathbf{s}_{1}}^{q}\right|du$,
$\dot{\mathbf{S}}_{\mathbf{s}_{1}}^{q}=\frac{d\mathbf{S}_{\mathbf{s}_{1}}^{q}}{du}$
(see Appendix \nameref{subsec:A.2-Specialization} / (\ref{eq:isotropic_inhomogeneous})).
The network ``energy'' $S_{net}$ is defined to be the sum of these
lengths:
\begin{equation}
S_{net}\coloneqq\oint_{q\in\mathbf{s}_{2}}\stackrel[u\left(q\right)]{u\left(\mathbf{s}_{1}\right)}{\int}\phi\left(\mathbf{S}_{\mathbf{s}_{1}}^{q}\right)\left|\dot{\mathbf{S}}_{\mathbf{s}_{1}}^{q}\right|dudv.\label{eq:net_energy}
\end{equation}
Minimizing $S_{net}$ is the equivalent of the task of finding all
minimal paths between the points $q=\mathbf{s}_{2}\left(v\right)$
and the $\mathbf{s}_{1}$ as zero set. (Note that the endpoints are
- by definition - dense on $\mathbf{s}_{2}$, but usually discontinuous
on $\mathbf{s}_{1}$.) Compared (\ref{eq:net_energy}) to the isotropic
minimal surface action \nameref{subsec:A.3-Minimal-Surface}/formula
(\ref{eq:functional_isotropic})
\begin{equation}
S=\stackrel[a]{u}{\int}\underset{v}{\oint}\Phi\left(\mathbf{S}\right)\left|\mathbf{S}_{v}\right|\left|\mathbf{S}_{u}\right|dvdu\label{eq:functional_isotropic-1}
\end{equation}
one can notice striking resemblance with the only real difference
being the presence of the multiplicator $\left|\mathbf{S}_{v}\right|$,
$\mathbf{S}_{v}=\frac{\partial\mathbf{S}}{\partial v}$ in the latter.
This quantity represents the infinitesimal distances $\left|d\mathbf{S}_{\left(v\right)}\right|=\left|\mathbf{S}_{v}\right|dv$
between the adjacent $u=const$ parameter lines. This slight difference
in the Lagrangians has huge impact in the result insomuch it inhibits
the parameter lines to diverge without control. 

Despite the semblance shown above, the nature of the real minimal
surface problem is fundamentally different as it requires minimal
path determination in the infinite dimensional function (triplet)
space: \nameref{subsec:A.3-Minimal-Surface}. In this paper we do
not attempt to search the solution in the infinite dimensional space.
Rather we incorporate \textit{divergence constraint} into the minimal
path Eikonal equation that - in some degree - imitates the effect
of the missing factor $\left|\mathbf{S}_{v}\right|$.

\section{Object Extraction with Modified Minimal Paths}

In this section first we collect some mathematical results that are
used in the proposed model then introduce the divergence constraint.

\subsection{Metric in the Vicinity of a Reference Surface }

In this section we turn our attention to the distance map solution
of the usual three dimensional Eikonal problem. The level sets of
this map are the locus of the points having constant distance defined
by a metric-based distance function, see \nameref{subsec:A.2-Specialization}).
We assume that these points - except isolated singularities - constitute
a regular surface in the space.

Spatial points $\mathbf{R}\in\mathbb{R}^{3}$ in the vicinity of a
reference surface $\mathbf{Q}\left(p,r\right)$ may be parameterized
as 
\begin{equation}
\mathbf{R}\left(p,r,s\right)=\mathbf{Q}\left(p,r\right)+s\mathbf{m}\left(p,r\right)
\end{equation}
where $\mathbf{m}\left(p,r\right)$ is the unit normal vector of the
surface at parameter values $\left(p,r\right)$; then $s$ is the
Euclidean distance measured from the surface. Spatial points $\mathbf{R}\left(p,r,s=const\right)$
constitute equidistant surfaces to $\mathbf{Q}\left(p,r\right)$.
The normal vectors of the equidistant surfaces are obtained by the
cross product of the coordinate (covariant) basis vectors $\mathbf{R}_{p}=\frac{\partial\mathbf{R}}{\partial p}$,
$\mathbf{R}_{r}=\frac{\partial\mathbf{R}}{\partial r}$: $\mathbf{M}\left(p,r,s\right)=\mathbf{R}_{p}\left(p,r,s\right)\times\mathbf{R}_{r}\left(p,r,s\right)$,
their magnitude can be calculated as $\left|\mathbf{M}\right|\left(p,r,s\right)=\mathbf{m}\left(p,r\right)\cdot\mathbf{M}\left(p,r,s\right)$,
\textit{i.e.
\begin{gather}
\left|\mathbf{M}\right|=\mathbf{m}\cdot\left[\mathbf{Q}_{p}\times\mathbf{Q}_{r}+s\left(\mathbf{Q}_{p}\times\mathbf{m}_{r}+\mathbf{m}_{p}\times\mathbf{Q}_{r}\right)+s^{2}\mathbf{m}_{p}\times\mathbf{m}_{r}\right].\label{eq:equid_surf_metric}
\end{gather}
}The coefficient of $s$ can be expressed with the inverse (contravariant)
basis $\mathbf{Q}^{p}$, $\mathbf{Q}^{r}$ as
\begin{gather*}
\mathbf{m}\cdot\left(\mathbf{Q}_{p}\times\mathbf{m}_{r}\right)=\mathbf{m}_{r}\cdot\left(\mathbf{m}\times\mathbf{Q}_{p}\right)=\left|\mathbf{Q}_{p}\times\mathbf{Q}_{r}\right|\mathbf{m}_{r}\cdot\mathbf{Q}^{r}\\
\mathbf{m}\cdot\left(\mathbf{m}_{p}\times\mathbf{Q}_{r}\right)=\mathbf{m}_{p}\cdot\left(\mathbf{Q}_{r}\times\mathbf{m}\right)=\left|\mathbf{Q}_{p}\times\mathbf{Q}_{r}\right|\mathbf{m}_{p}\cdot\mathbf{Q}^{p}
\end{gather*}
(easy to check that indeed: $\mathbf{Q}_{i}\cdot\mathbf{Q}^{k}=\delta_{i}^{k}$,
$i,k\in\left\{ p,r\right\} $ where $\delta_{i}^{k}$ is the Kronecker
delta). Denoting the reference surface normal by $\mathbf{M}_{0}\left(=\mathbf{M}\left(r,s,0\right)\right)=\mathbf{Q}_{p}\times\mathbf{Q}_{r}$,
expression (\ref{eq:equid_surf_metric}) becomes:\textit{
\begin{gather}
\left|\mathbf{M}\right|=\left|\mathbf{M}_{0}\right|\left[1+s\left(\mathbf{m}_{p}\cdot\mathbf{Q}^{p}+\mathbf{m}_{r}\cdot\mathbf{Q}^{r}\right)\right]+s^{2}\mathbf{m}\cdot\left(\mathbf{m}_{p}\times\mathbf{m}_{r}\right).\label{eq:equid_surf_metric-1}
\end{gather}
}In (\ref{eq:equid_surf_metric-1}) $\mathbf{m}_{p}\cdot\mathbf{Q}^{p}+\mathbf{m}_{r}\cdot\mathbf{Q}^{r}$
is the divergence of the unit normal vector $\nabla\cdot\mathbf{m}$,
known to be the negative of the sum curvature $K_{S}$. (Also, by
simple calculation, the last term is $s^{2}\mathbf{m}\cdot\left(\mathbf{m}_{p}\times\mathbf{m}_{r}\right)=s^{2}\left|\mathbf{M}_{0}\right|K_{G}$
with $K_{G}$ being the Gaussian curvature.)

Considering non-equidistant surfaces, the direction of the unit normal
vector is the function of $s$: $\mathbf{m}\left(p,r,s\right)\neq\mathbf{m}\left(p,r\right)$.
In this case, equation (\ref{eq:equid_surf_metric-1}) is applicable
only to the immediate neighbors with infinitesimal distance $ds$
from $\mathbf{Q}$, then the last term $ds^{2}K_{G}$ becomes insignificant,
hence omitted:
\begin{alignat}{1}
\left|\mathbf{M}\right|\left(p,r,ds\right) & =\left|\mathbf{M}_{0}\right|\left[1+\left(\nabla\cdot\mathbf{m}\right)ds\right]\nonumber \\
 & =\left|\mathbf{M}_{0}\right|\left(1-K_{S}ds\right).\label{eq:immediate_metric}
\end{alignat}

Same holds for the invariant elementary surface area $dA=\left|\mathbf{M}\right|dpdr$. 

\subsection{Alteration of the Area Element of an Evolving Surface}

\label{subsec:Alteration-of-the-area}

Now, we track the \textit{change of the magnitude} of a surface patch
evolving in the normal direction (that is $p,\,r$ are constant).
Here we identify the reference surface with the $S$ level set of
an evolving surface and assume that the rate of change of the Euclidean
dilation wrt the level set value $\frac{ds}{dS}\left(S\right)$ is
known.\footnote{In the case, the evolving surface is the front of a distance map under
construction, the level set values are the geodesic distances measured
from the zero set.} From (\ref{eq:immediate_metric}) we obtain
\begin{gather}
\left|\mathbf{M}\right|\left(S+dS\right)=\left|\mathbf{M}\right|\left(S\right)\left(1-K_{S}\left(S\right)\frac{ds\left(S\right)}{dS}dS\right).\label{eq:diff_area_change}
\end{gather}
(\ref{eq:diff_area_change}) leads to the differential equation $\frac{d\left|\mathbf{M}\right|}{dS}\left(S\right)=-\left|\mathbf{M}\right|\left(S\right)K_{S}\left(S\right)\frac{ds}{dS}$
with the solution\footnote{Level set value 0 is assumed on the boundary of the initial action
map.} 
\begin{equation}
\left|\mathbf{M}\right|\left(S\right)=\left|\mathbf{M}\right|\left(0\right)\exp\left(-\stackrel[0]{S}{\int}K_{S}\left(\hat{S}\right)\frac{ds}{d\hat{S}}d\hat{S}\right);
\end{equation}
however, we will not need this cumulative form, for the reason that
we always consider unit area around each point of the ever growing
minimal action map boundary. \begin{figure}[!htbp]  
 \centering
 \includegraphics[width=0.33\columnwidth]{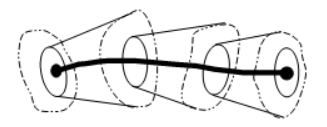}
 \caption{A path with the evolving unit area around.} 
\label{fig:pathtube}
\end{figure}

The area of an elementary surface patch also changes with the rate
of (\ref{eq:diff_area_change}). The alteration of the unit area is
\textbf{(Fig.}\ref{fig:pathtube}): 
\begin{gather}
\frac{\left|\mathbf{M}\right|\left(S+dS\right)}{\left|\mathbf{M}\right|\left(S\right)}=\left(1-K_{S}\left(S\right)\frac{ds\left(S\right)}{dS}dS\right).\label{eq:diff_area_change-1}
\end{gather}
The length of the one-dimensional structures inscribed in an elementary
surface patch is - in some average sense - proportional with the square
root of the area of that patch. The real rate of change of the elementary
lengths (vector magnitudes) is the function of the direction in the
tangent plane that the one-dimensional structure passes. If we wish
to incorporate some elementary length/local distance information into
our equations, but also want to maintain their isotropic nature we
are left with this ``in average sense'' possibility. Therefore,
the change rate of the ``mean elementary distances'' can be expressed
as $\sqrt{\frac{\left|\mathbf{M}\right|\left(S+dS\right)}{\left|\mathbf{M}\right|\left(S\right)}}=\sqrt{1-K_{S}\left(S\right)\frac{ds\left(S\right)}{dS}dS}$
that for small sum curvatures can further be approximated with $1-H\frac{ds\left(S\right)}{dS}dS$,
where $H=\frac{1}{2}K_{S}$ is the mean curvature. 

Based on these reasonings, now we can turn our attention to the divergence
constraint.

\subsection{The Divergence Constraint}

\label{subsec:The-Divergence-Constraint}

The distance map solutions for the minimal path Eikonal problems can
be obtained by the updating scheme \nameref{subsec:A.2-Specialization}
/(\ref{eq:Eikonal-1}) 
\begin{equation}
\delta S=\mathbf{p}\cdot\delta\mathbf{r}\label{eq:Eikonal-1-1}
\end{equation}
gradually expanding the initial zero set - where the action value
$S$ is zero - in the normal direction. The boundary of the momentary
map (the ``front'' surface) always represents a constant level set
$S=const$ of the distance map. Raising the action value from $S$
to $S+\delta S$, the dilation of the map $\left|\delta\mathbf{r}\right|$
in the normal direction is obtained for known $\left|\mathbf{p}\right|$.\footnote{In the isotropic case, the value of $\mathbf{\left|p\right|}$ is
completely determined by the image content.} The momentum $\mathbf{p}$ then can be retrieved from the constructed
distance map in the gradient descent direction. In general, the tangent
of the minimal path is not the homogeneous function of $\mathbf{p}$,
but it is in the isotropic case.

Summary of the isotropic case is as follows. The functional is defined
as
\begin{equation}
S=\stackrel[a]{t}{\int}\phi\left(\mathbf{r}\right)\left|\dot{\mathbf{r}}\right|d\hat{t}\label{eq:action_inhomo_isotrop}
\end{equation}
and the corresponding Eikonal equation $\nabla S=\mathbf{p}$ $\left(\mathbf{p}=\frac{\partial\phi\left|\dot{\mathbf{r}}\right|}{\partial\dot{\mathbf{r}}}\right)$
becomes
\begin{equation}
\nabla S=\phi\left(\mathbf{r}\right)\mathbf{e},
\end{equation}
where $\mathbf{e}=\frac{\dot{\mathbf{r}}}{\left|\dot{\mathbf{r}}\right|}$
is the unit tangent of the minimal paths, \textbf{coinciding} with
the unit normal $\mathbf{m}$ of the level sets of the distance map
at the same point in $\mathbb{R}^{3}$. The updation scheme in the
normal direction of the frontal $S=const$ level set is $\delta S=\left|\mathbf{p}\right|\left|\delta\mathbf{r}\right|$
where dilation $\left|\delta\mathbf{r}\right|$ is (also) the elongation
of the minimal paths. Hereinafter we denote it by $\delta s\coloneqq\left|\delta\mathbf{r}\right|$
(also index $0$ is attached to the variation of the action value
for later referencing):
\begin{gather}
\delta s=\frac{\delta S_{0}}{\phi\left(\mathbf{r}\right)}.\label{eq:updation_scheme_isotropic}
\end{gather}

We wish to incorporate a quantity into (\ref{eq:updation_scheme_isotropic})
that reduces the unit area/mean distance between the adjacent minimal
paths. By (\ref{eq:immediate_metric}),(\ref{eq:diff_area_change-1})
this task is equivalent to the reduction of the divergence of the
unit normal $\mathbf{m}\left(=\mathbf{e}\right)$ of the level sets.\footnote{The reduction of the divergence can be achieved in many ways. We look
for a direct modification of the updation sceme (\ref{eq:updation_scheme_isotropic}).} Lets assume, that the front surface is the level set of the action
map at some value $S$. We consider the \textit{modified action} increment,
corrected by either the sum curvature (\ref{eq:diff_area_change-1})
or the mean curvature (hereinafter $K\left(S\right)$ stands for either
one of them) examined in the previous section \nameref{subsec:Alteration-of-the-area}
\begin{equation}
\delta S\coloneqq\delta S_{0}\left(1-\lambda K\left(S\right)\frac{ds\left(S\right)}{dS}\delta S_{0}\right),\:\lambda\in\left[0,1\right].
\end{equation}
From (\ref{eq:action_inhomo_isotrop}) $\left(ds=\left|\dot{\mathbf{r}}\right|dt\rightarrow\right)$
$\frac{ds}{dS}=\frac{1}{\phi}$. Also substituting the variation of
the action value (\ref{eq:updation_scheme_isotropic}) $\delta S_{0}=\phi\delta s$
and choosing $\lambda=1$ to maximize the effect of our \textit{divergence
constraint}, we get the following quadratic equation
\begin{gather}
\delta S=\phi\delta s\left(1-K\delta s\right)\rightarrow\nonumber \\
K\delta s^{2}-\delta s+\frac{\delta S}{\phi}=0.\label{eq:modified_equation}
\end{gather}
This is a second-order correction to (\ref{eq:updation_scheme_isotropic})
as it includes the square of the sought normal dilation $\delta s$
- the value equivalent to the elongation of the minimal path. The
solution for this value is:
\begin{equation}
\delta s=\frac{1-\sqrt{1-4K\frac{\delta S}{\phi}}}{2K}.\label{eq:modified_update}
\end{equation}
The sign of the square root is determined by the sign convention chosen
for the curvatures and the limit value $\underset{K\rightarrow0}{\lim}\delta s$,
since in this limit the solution of (\ref{eq:modified_equation})
needs to be reduced to the original (\ref{eq:updation_scheme_isotropic}).

\subsection{Discussing the Divergence Constraint}

In this section we refer formula (\ref{eq:modified_update}) as $\delta s_{corr}$
and assume same $\delta S$ elevation in the in the action value for
both update schemes \ref{eq:updation_scheme_isotropic}-\ref{eq:modified_update}.
First we check the limit of $\delta s_{corr}$ for zero curvatures.
By using L'Hopital's rule, it is 
\begin{equation}
\underset{K\rightarrow0}{\lim}\frac{1-\sqrt{1-4K\frac{\delta S}{\phi}}}{2K}=\underset{K\rightarrow0}{\lim}-\frac{-4\frac{\delta S}{\phi}}{4\sqrt{1-4K\frac{\delta\widetilde{S}}{\phi}}}=\frac{\delta S}{\phi}
\end{equation}
 as expected.

Formula (\ref{eq:modified_update}) defines the distance map dilation
$\delta s$ in the normal direction as the function of the geodesic
distance elevation $\delta S$ and the curvature $K$ values. We used
the sign convention $K_{S}=-\nabla\cdot\mathbf{m}$. Hence, the divergence
of the outward unit vectors are a) positive for the points where the
action map stretches forward and b) negative for the points lagging
behind their neighbors. For the same elevation of the action values
the corresponding advances: $\delta s_{corr}<\delta s$ in case a)
and $\delta s_{corr}>\delta s$ in case b), causing more uniform evolution
of the action map. This in turn reduces the divergence of the normal
vectors on the map boundary - or equivalently the divergence of the
minimal paths (recall, in the isotropic case, $\dot{\mathbf{r}}$
is the homogeneous function of the surface normal: $\dot{\mathbf{r}}\varpropto\mathbf{m}$
).

For extremely high - positive, case b) - curvatures, the discriminant
$1-4K\frac{\delta S}{\phi}$ can become negative. To avoid this situation,
the maximum elevation of the action value needs to be limited to 
\begin{equation}
\delta S\leq\frac{\phi}{4K},\label{eq:action_limit}
\end{equation}
where the right hand side is the worst case among the points on the
momentary distance map boundary. 

Another updation strategy can be to choose the constant action value
elevation $\delta S\coloneqq\phi_{min}\delta s_{max}$. Here $\phi_{min}$
is the smallest possible value - determined by the image and $\delta s_{max}$
is a constant parameter - the maximum enabled normal-dilation increment
by \ref{eq:updation_scheme_isotropic}. Then wherever the discriminant
would become negative, we have to limit the dilation to the maximum
value $\delta s=2\frac{\delta S}{\phi}$.

\section{Results}

\begin{figure}[!htbp]
 \centering
 \begin{tabular}{c c c}
	\includegraphics[width=0.4\columnwidth]{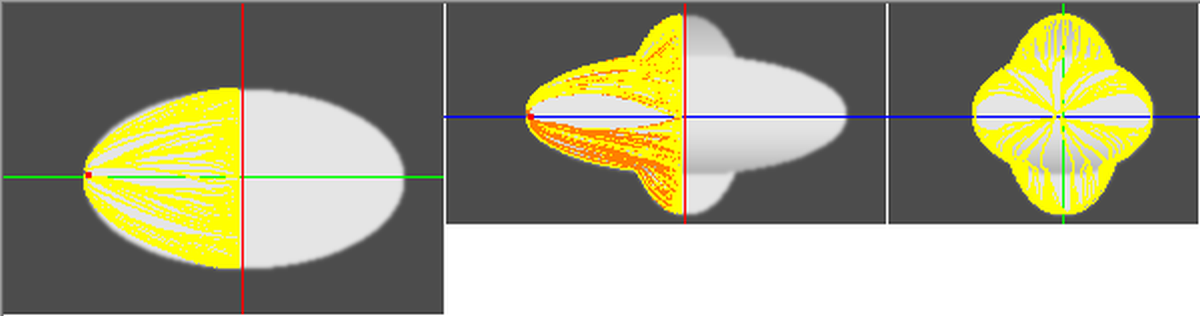}&
	\includegraphics[width=0.4\columnwidth]{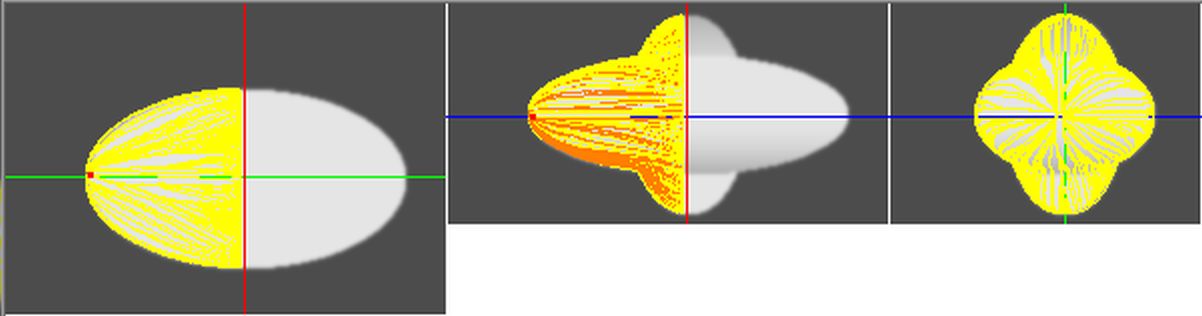}\\
	
	\includegraphics[width=0.4\columnwidth]{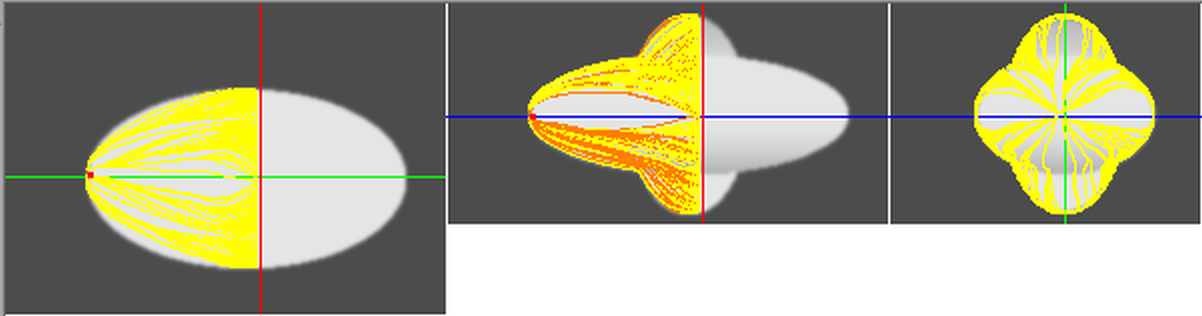}&
	\includegraphics[width=0.4\columnwidth]{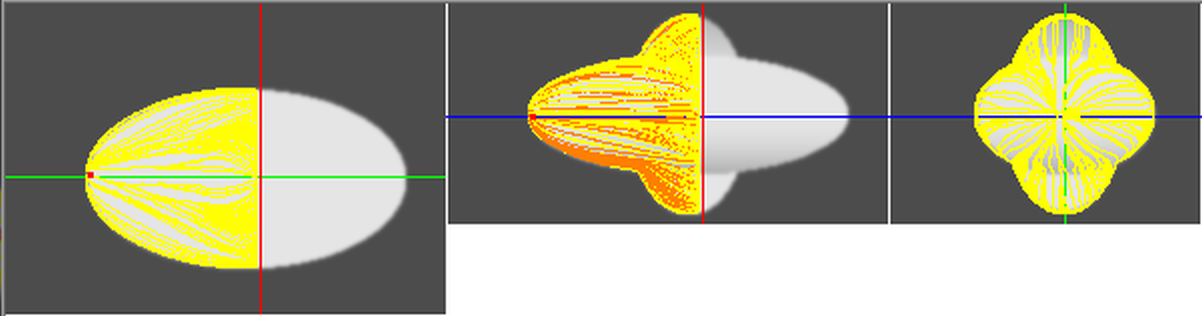}\\

	\includegraphics[width=0.4\columnwidth]{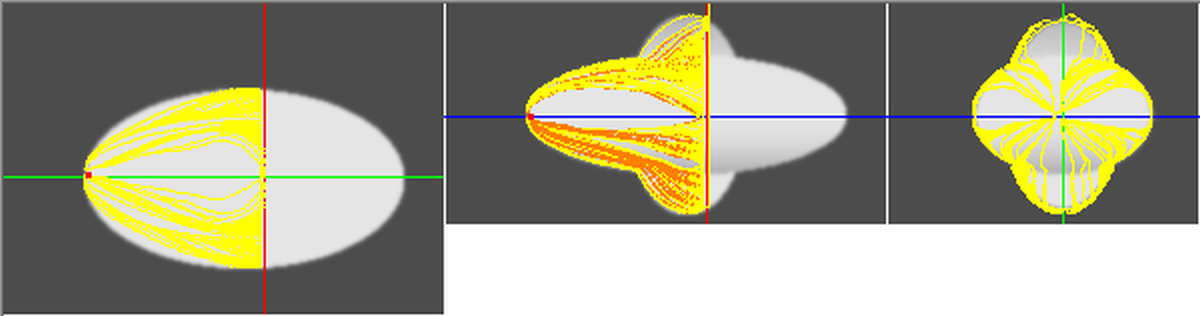}&
	\includegraphics[width=0.4\columnwidth]{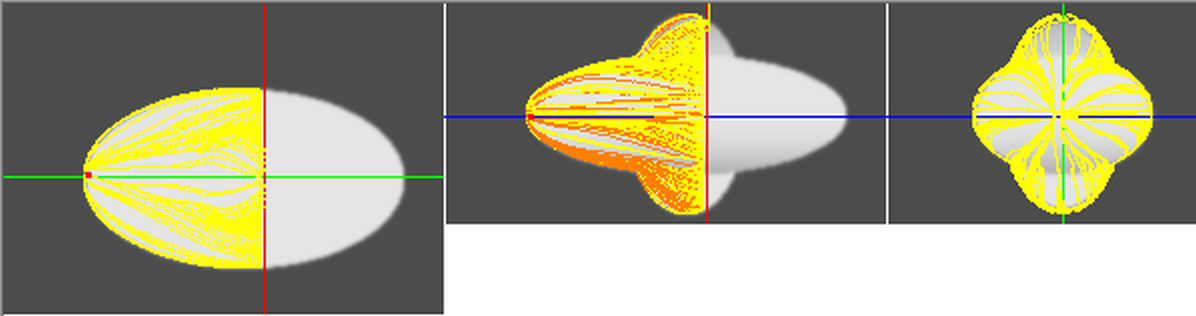}\\

 \end{tabular}
 \caption{ Columns: The top-side-front views of the path networks calculated using the original (left) and the divergence-corrected (right) distance map updation schemes. Rows: results obtained from increasingly asymmetrical plane setup. As the asymmetry increased more surface area becomes deserted. }
\label{fig:result1}
\end{figure}

The method was tested on a synthetic voxel intensity image. The test
environment and the user inputs are as follows.

The test object is composed from two inhomogeneous ellipsoids. Artificial
noise is added. The user input is a point and a plane. The initial
distance map (the zero set) is an $\varepsilon$-ball around the point
with the radius of 3-voxels. The map evolves until all pixels of the
plane are reached. Then the object contour on the plane ($\mathbf{s}_{2}$
in \nameref{sec:The-Minimal-Path-net}) is automatically detected.
Finally, the path networks are retrieved starting from the contour
of the object-plane intersection. The comparative results, using update
schemes a) (\ref{eq:updation_scheme_isotropic}):
\begin{gather*}
\delta s=\frac{\delta S}{\phi}
\end{gather*}
and b) (\ref{eq:modified_update}):
\[
\delta s=\frac{1-\sqrt{1-4K\frac{\delta S}{\phi}}}{2K},
\]
for increasingly asymmetrical plane setup, are shown in \textbf{Fig.}\ref{fig:result1}.
The enforced space metric is an isotropic inhomogeneous metric extracted
from the image
\begin{equation}
\phi\left(\mathbf{r}\right)=\alpha+\left(1-\alpha\exp\left(-\beta\left|\nabla I\left(\mathbf{r}\right)\right|\right)\right),
\end{equation}
with the choices $\alpha=0.01$, $\beta=14$. $\left|\nabla I\left(\mathbf{r}\right)\right|$
is the magnitude of the gradient of the normed ($I\in\left[0,1\right]$)
image intensity function.

\section{Conclusion}

In this paper we presented a second order correction to the inhomogeneous
minimal path Eikonal equations that prevents the adjacent minimal
path trajectories to diverge uncontrollably. We showed that the minimal
path network obtained from the modified equations greatly reduces
the surface area left uncovered with paths. This denser covering greatly
improves the accuracy of the approximate surface that can be retrieved
directly from the path network. It can also be used to initialize
the transport equation enhancing its robustness.

\section*{Appendix: The minimal surface problem}

\label{sec:Appendix}

In the appendix we 1) summarize the relevant aspects of the Hamilton-Jacobi
Theorem; 2) Its specialization to the Minimal Path Eikonal equation;
2) The generalization of the Minimal Path Eikonal equation to the
Minimal Surface problem.

\subsection*{A.1 The Hamilton-Jacobi equation}

The functional of variable endpoints $S\left(q_{i},t\right)$ of $n$
function $q_{i}\left(t\right)$, $i=1\ldots n$: 
\begin{equation}
S\left(q_{i},t\right)=\stackrel[a]{t}{\int}L\left(q_{i},\dot{q}_{i},\hat{t}\right)d\hat{t}
\end{equation}
 can always be decomposed into the sum
\begin{equation}
S\left(q_{i},t\right)=\stackrel[i=1]{n}{\sum}\stackrel[q_{a}]{q_{t}}{\int}p_{i}d\hat{q}_{i}-\stackrel[a]{t}{\int}H\left(q_{i},p_{i},\hat{t}\right)d\hat{t}\label{eq:action_decomp}
\end{equation}
where $p_{i}=\frac{\partial L}{\partial q_{i}}$, $i=1\ldots n$ are
the generalized momenta, $H\left(q_{i},p_{i},t\right)=\stackrel[i=1]{n}{\sum}p_{i}\dot{q}_{i}-L$
is the Hamiltonian of the problem. Decomposition (\ref{eq:action_decomp})
arises from the analysis of the partial alterations of the action
value $S$ as the function of the change in the boundary conditions
wrt a) the end coordinates (the $1^{\mathrm{st}}$ terms on the right)
and b) the parameter value at the end coordinates (the $2^{\mathrm{nd}}$
term on the right). The corresponding local equations are:
\begin{align}
\mathrm{a}) & \:p_{i}=\frac{\partial S}{\partial q_{i}},\:i=1\ldots n\nonumber \\
\mathrm{b}) & \:\frac{\partial S}{\partial t}=-H,\:H=H\left(q_{i},p_{i},t\right).\label{eq:Hamilton_partials}
\end{align}
Combining a) and b), one can get the Hamilton-Jacobi PDE:
\begin{equation}
\frac{\partial S}{\partial t}+H\left(q_{i},\frac{\partial S}{\partial q_{i}},t\right)=0\label{eq:Hamilton-Jacobi}
\end{equation}
expressing the relation between partial variations a) and b).

\subsection*{A.2 Specialization}

\label{subsec:A.2-Specialization}

We restrict our further examination to the geometric functionals.
By definition, a functional is geometric, if it is immune to the reparameterization,
that is, for any admissible $t\rightarrow\gamma\left(t\right)$ we
require:
\begin{equation}
\int L\left(q_{i},\dot{q}_{i},t\right)dt=\int L\left(q_{i},\frac{dq_{i}}{d\gamma}\dot{\gamma},t\left(\gamma\right)\right)\frac{1}{\dot{\gamma}}d\gamma\coloneqq\int L\left(q_{i},\frac{dq_{i}}{d\gamma},\gamma\right)d\gamma.\label{eq:geom_defined}
\end{equation}
Because of this independence criterion, in the geometric case, only
the change of coordinates -- encoded in the $1^{\mathrm{st}}$ terms
on the right of (\ref{eq:action_decomp}) -- has effect, hence requirement
(\ref{eq:geom_defined}) implies identically zero Hamiltonian.\footnote{This further implies scleronomic $\left(\frac{\partial L}{\partial t}=0\right)$functional.}
Then formula (\ref{eq:action_decomp}) is reduced to the abbreviated
functional
\begin{gather}
S\left(q_{i}\right)=\stackrel[i=1]{n}{\sum}\stackrel[q_{a}]{q_{t}}{\int}p_{i}d\hat{q}_{i},\nonumber \\
p_{i}=\frac{\partial L}{\partial q_{i}},i=1\ldots n\label{eq:action_abbreviated}
\end{gather}
describing static action map. 

Henceforth the general coordinates $q_{i}$ are identified with the
coordinate functions of a \textquotedblright real\textquotedblright{}
position vector $\mathbf{r}\left(t\right)\in\mathbb{R}^{3}$, (or
for planar problems $\mathbf{r}\left(t\right)\in\mathbb{R}^{2}$),
the Hamilton-Jacobi equation (\ref{eq:Hamilton-Jacobi}) becomes the
Eikonal equation that can concisely be written
\begin{equation}
\frac{dS}{d\mathbf{r}}\left(=\nabla S\right)=\mathbf{p},\:\mathbf{p}=\frac{\partial L}{\partial\dot{\mathbf{r}}}.\label{eq:Eikonal}
\end{equation}
(From (\ref{eq:Hamilton_partials}) b) $\frac{\partial S}{\partial t}=0$
$\rightarrow$ $\frac{\partial S}{\partial\mathbf{r}}=\frac{dS}{d\mathbf{r}}$,
hence a) $\frac{dS}{d\mathbf{r}}=\mathbf{p}$). The equivalent formulation
using differentials 
\begin{equation}
\delta S=\mathbf{p}\cdot\delta\mathbf{r}\label{eq:Eikonal-1}
\end{equation}
is used to construct the solution by gradually extending the action
map starting from a zero set (often chosen to be an ``$\varepsilon$-ball'').

The static action map solution of the Hamilton-Jacobi equation for
the geometric functionals can be identified with \textit{distance
map} wrt a distance function defined by metric, where the metric of
the space is encoded in the Lagrangian. The parameterization-independent
Finsler metric:
\begin{gather}
L=\psi\left(\mathbf{r},\mathbf{e}\right)\left|\dot{\mathbf{r}}\right|\nonumber \\
\mathbf{e}=\frac{\dot{\mathbf{r}}}{\left|\dot{\mathbf{r}}\right|},\label{eq:Finsler}
\end{gather}
realizes the most general form of such functionals. Its specialization
sequence includes the asymmetric Randers metric:
\begin{gather}
L=\left(\sqrt{\mathbf{e}\cdot\mathbf{G}\cdot\mathbf{e}}+\boldsymbol{\omega}\cdot\mathbf{e}\right)\left|\dot{\mathbf{r}}\right|,
\end{gather}
the anisotropic Riemannian metric ($\mathbf{G}$ stands for the positive
definite symmetric metric tensor):
\begin{equation}
L=\sqrt{\mathbf{e}\cdot\mathbf{G}\cdot\mathbf{e}}\left|\dot{\mathbf{r}}\right|\:\left(=\sqrt{\dot{\mathbf{r}}\cdot\mathbf{G}\cdot\dot{\mathbf{r}}}\right),\label{eq:Riemann_length}
\end{equation}
the inhomogeneous, isotropic metric:
\begin{equation}
L=\phi\left(\mathbf{r}\right)\left|\dot{\mathbf{r}}\right|\:\left(\mathbf{G}=\phi^{2}\left(\mathbf{r}\right)\mathbf{I}\right),\label{eq:isotropic_inhomogeneous}
\end{equation}
and the Euclidean metric $\left(\phi\left(\mathbf{r}\right)\equiv const\right)$.
The ratio of the Riemannian and Euclidean lengths $\frac{\sqrt{\dot{\mathbf{r}}\cdot\mathbf{G}\cdot\dot{\mathbf{r}}}}{\left|\dot{\mathbf{r}}\right|}=\sqrt{\mathbf{e}\cdot\mathbf{G}\cdot\mathbf{e}}$,
$\left|\mathbf{e}\right|=1$ is encoded in the one-dimensional minor
of the metric tensor in the direction of the path tangent.

In the isotropic case (and only in the isotropic case), the minimal
paths are perpendicular to the level sets of the of the distance map:
$\mathbf{p}=\frac{\partial L}{\partial\dot{\mathbf{r}}}=\phi\left(\mathbf{r}\right)\mathbf{e},$
hence the abbreviated functional is obtained by rearrangement: $\int\phi\left(\mathbf{r}\right)\left|\dot{\mathbf{r}}\right|dt=$$\int\phi\left(\mathbf{r}\right)\left(\mathbf{e}\cdot\dot{\mathbf{r}}\right)dt=$$\int\phi\left(\mathbf{r}\right)\mathbf{e}\cdot\left(\dot{\mathbf{r}}dt\right)=$$\int\left(\phi\left(\mathbf{r}\right)\mathbf{e}\right)\cdot d\mathbf{r}$.

\subsection*{A.3 The Minimal Surface problem\label{subsec:A.3-Minimal-Surface}}

\begin{figure}[!htbp]
 \centering
 \includegraphics[width=0.5\columnwidth]{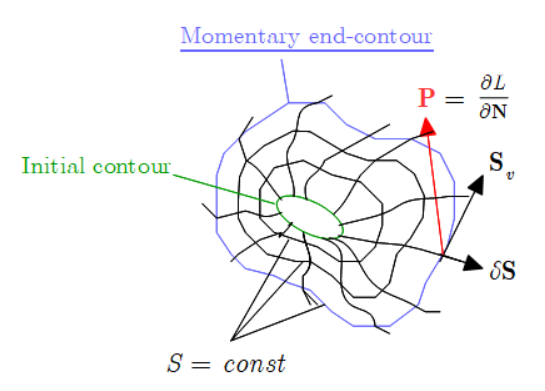}
 \caption { One particular path. S = const contours are boundary points on the consecutive action (hyper-)surface. }
\label{fig:onepath}
\end{figure}

Summarizing the discussion of the previous sections, the functionals
$\int Ldt$ for the one-parameter minimal path problems can be reformulated
as $\int\mathbf{p}\cdot d\mathbf{r}$, where $\mathbf{p}=\frac{\partial L}{\partial\dot{\mathbf{r}}}$
is the new (momentum) variable and the invariant integration variable
is the tangent times the path parameter $d\mathbf{r}=\dot{\mathbf{r}}dt$.
We seek analogous formulation -- including the corresponding Eikonal
equations -- for the two-parameter minimal surface problem. The minimal
path is the shortest distance path among all possible paths joining
point pairs in $\mathbb{R}^{3}$. Similarly, the ``shortest distance
between two contours'' can be identified by the surface having the
the minimal surface area among all possible surfaces connecting them.

\textbf{Notation}: The two-parameter position vector to the surface
points is denoted by $\mathbf{S}\left(u,v\right)$. Let the directed
invariant surface element be denoted by $d\mathbf{A}=\mathbf{N}dudv$
-- where $\mathbf{N}\left(u,v\right)$ is the normal of the surface
at parameter values $\left(u,v\right)$ -- it can be expressed by
the local covariant basis vectors $\mathbf{S}_{u}=\frac{\partial\mathbf{S}}{\partial u}$,
$\mathbf{S}_{v}=\frac{\partial\mathbf{S}}{\partial v}$ as $\mathbf{N}=\mathbf{S}_{u}\times\mathbf{S}_{v}$;
let the unit normal vector be $\mathbf{n}=\frac{\mathbf{N}}{\left|\mathbf{N}\right|}$.
The absolute value (magnitude) of the directed invariant surface element
expresses the elementary Euclidean surface area: $\left|\mathbf{N}\right|dudv\coloneqq dA$. 

The general form of the functional given by double integrals with
Lagrangian having only the first order partial derivatives is:
\begin{equation}
\iint_{A}L\left(\mathbf{S},\mathbf{S}_{u},\mathbf{S}_{v},u,v\right)dudv.\label{eq:general_doubleint}
\end{equation}
We restrict our further examination to the geometric case. The two
dimensional equivalent of the geometric Finsler functional (\ref{eq:Finsler})
is:
\begin{gather}
S=\iint_{A}\Psi\left(\mathbf{S},\mathbf{n}\right)dA.\label{eq:Surf_Finsler}
\end{gather}
with Lagrangian $L=\Psi\left(\mathbf{S},\mathbf{n}\right)\left|\mathbf{N}\right|$.
Note that it does not include the covariant basis vectors (neither
the parameters) explicitly. Hereinafter we \textbf{assume} that the
Lagrangian of the minimal surface problem must not include more explicitness
and work by analogy. One may define the\textit{ surface momentum}
\begin{equation}
\mathbf{P}\coloneqq\frac{\partial L}{\partial\mathbf{N}}\label{eq:Surf_moment}
\end{equation}
and the abbreviated functional as
\begin{equation}
S=\iint_{A}\mathbf{P}\cdot d\mathbf{A}.\label{eq:Surf_Finsler_abbr}
\end{equation}
In the case of the Finsler functional (\ref{eq:Surf_Finsler}), $\mathbf{P}=\Psi_{\mathbf{n}}\cdot\left(\mathbf{I}-\mathbf{nn}\right)+\Psi\mathbf{n}$$=\left.\Psi_{\mathbf{n}}\right|_{T}+\Psi\mathbf{n}$
- where $\mathbf{I}$ is the identity tensor ($\mathbf{I}-\mathbf{nn}$
is the projector to the tangent space, $\left.\cdot\right|_{T}$ stands
for the projection to the tangent space) and $\Psi_{\mathbf{n}}=\frac{\partial\Psi}{\partial\mathbf{n}}$.
Then (\ref{eq:Surf_Finsler_abbr}) indeed gives (\ref{eq:Surf_Finsler})
back.

In agreement with our explicitness assumption, now we consider the
effect of the varying boundary condition for the particular form of
the double double integral
\begin{equation}
S=\iint L\left(\mathbf{S},\mathbf{N}\right)dudv.\label{eq:func_to_vary}
\end{equation}
This form is more general than (\ref{eq:Surf_Finsler}) but does not
depend explicitly on the covariant basis $\mathbf{S}_{u}$, $\mathbf{S}_{v}$.
First, we need the variation of the surface normal in the direction
of $\mathbf{h}_{\mathbf{N}}$:
\begin{gather}
\mathbf{N}+\varepsilon\mathbf{h}_{\mathbf{N}}=\left(\mathbf{S}_{u}+\varepsilon\mathbf{h}_{u}\right)\times\left(\mathbf{S}_{v}+\varepsilon\mathbf{h}_{v}\right)\qquad\qquad\qquad\nonumber \\
=\mathbf{N}+\varepsilon\left(\mathbf{h}_{u}\times\mathbf{S}_{v}+\mathbf{S}_{u}\times\mathbf{h}_{v}\right)\:\left(+\varepsilon^{2}\mathbf{h}_{u}\times\mathbf{h}_{v}\right)\\
\rightarrow\:\delta\mathbf{N}=\varepsilon\left(\mathbf{h}_{u}\times\mathbf{S}_{v}+\mathbf{S}_{u}\times\mathbf{h}_{v}\right)\nonumber 
\end{gather}
then the variation of (\ref{eq:func_to_vary}) is
\begin{gather}
\delta S=\iint L\left(\mathbf{S}+\varepsilon\mathbf{h},\mathbf{N}+\varepsilon\mathbf{h}_{\mathbf{N}}\right)-L\left(\mathbf{S},\mathbf{N}\right)dudv\qquad\qquad\qquad\qquad\nonumber \\
=\iint\frac{\partial L}{\partial\mathbf{S}}\cdot\varepsilon\mathbf{h}+\frac{\partial L}{\partial\mathbf{N}}\cdot\left(\varepsilon\mathbf{h}_{u}\times\mathbf{S}_{v}+\mathbf{S}_{u}\times\varepsilon\mathbf{h}_{v}\right)dudv\qquad\qquad\\
=\iint\varepsilon\mathbf{h}\cdot\frac{\partial L}{\partial\mathbf{S}}+\varepsilon\mathbf{h}_{u}\cdot\left(\mathbf{S}_{v}\times\frac{\partial L}{\partial\mathbf{N}}\right)+\varepsilon\mathbf{h}_{v}\cdot\left(\frac{\partial L}{\partial\mathbf{N}}\times\mathbf{S}_{u}\right)dudv.\nonumber 
\end{gather}
After applying the integration by parts step, we take the usual assumption
that at the extremal surface the Euler-Lagrange term vanishes. Also,
the boundary terms are reduced to boundary integrals by the Green's
theorem, then we are left with (here the notations $d\mathbf{S}_{\left(u\right)}=\mathbf{S}_{u}du$,
$d\mathbf{S}_{\left(v\right)}=\mathbf{S}_{v}dv$ and the ``cross
tensor'': $\mathbf{v}\cdot\left[\mathbf{p}\right]_{\times}\equiv\mathbf{v}\times\mathbf{p}$,
$\left[\mathbf{p}\right]_{\times}\cdot\mathbf{w}\equiv\mathbf{p}\times\mathbf{w}$,
$\forall\:\mathbf{v},\mathbf{w},\mathbf{p}\in\mathbb{R}^{3}$ are
used):
\begin{gather}
\delta S=0+\iint\frac{\partial}{\partial u}\left(d\mathbf{S}_{\left(v\right)}\cdot\left[\frac{\partial L}{\partial\mathbf{N}}\right]_{\times}\cdot\varepsilon\mathbf{h}\right)du+\iint\frac{\partial}{\partial v}\left(\varepsilon\mathbf{h}\cdot\left[\frac{\partial L}{\partial\mathbf{N}}\right]_{\times}\cdot d\mathbf{S}_{\left(u\right)}\right)dv\nonumber \\
=\oint d\mathbf{S}_{\left(v\right)}\cdot\left[\frac{\partial L}{\partial\mathbf{N}}\right]_{\times}\cdot\varepsilon\mathbf{h}+d\mathbf{S}_{\left(u\right)}\cdot\left[\frac{\partial L}{\partial\mathbf{N}}\right]_{\times}\cdot\varepsilon\mathbf{h}\qquad\qquad\qquad\\
=\oint\left(d\mathbf{S}_{\left(v\right)}+d\mathbf{S}_{\left(u\right)}\right)\cdot\left[\frac{\partial L}{\partial\mathbf{N}}\right]_{\times}\cdot\varepsilon\mathbf{h}.\nonumber 
\end{gather}
Note that in the case of geometric functionals this is the only non-zero
term, arising alone from the boundary variation. Finally, using the
usual notation $\varepsilon\mathbf{h}=\delta\mathbf{S}$, also the
fact that $d\mathbf{S}_{\left(v\right)}+d\mathbf{S}_{\left(u\right)}$
is tangent to the boundary curve (hence denoted by $d\mathbf{S}_{B}$)
we arrived to the differential form of the Eikonal equation - an equivalent
to (\ref{eq:Eikonal-1}) - for minimal surfaces:
\begin{equation}
\delta S=\oint_{\delta A}d\mathbf{S}_{B}\cdot\left[\frac{\partial L}{\partial\mathbf{N}}\right]_{\times}\cdot\delta\mathbf{S}
\end{equation}
where $\delta A$ stands for the varying end of the surface. Now it
is tempting to choose special parameterization to get deeper insight
into the minimal surface problem. Adapting the $u=const$ parameter
lines to the boundary of the momentary surface -- that is with identifications
$d\mathbf{S}_{B}\rightarrow d\mathbf{S}_{\left(v\right)}$, $d\mathbf{S}_{\left(v\right)}=\mathbf{S}_{v}dv$
and using the surface moment (\ref{eq:Surf_moment}):
\begin{equation}
\delta S=\underset{v}{\oint}dv\left(\mathbf{S}_{v}\times\mathbf{P}\right)\cdot\delta\mathbf{S}.\label{eq:Eikonal_surf_difform}
\end{equation}
(Note: integrating (\ref{eq:Eikonal_surf_difform}): $\underset{u}{\int}\delta Sdu$
we get back the abbreviated Finsler functional (\ref{eq:Surf_Finsler_abbr})
parameterized specially: $\stackrel[a]{u}{\int}\underset{v}{\oint}\frac{\partial L}{\partial\mathbf{N}}\cdot\mathbf{N}dudv$.)
By analogy (see (\ref{eq:Eikonal}), (\ref{eq:Eikonal-1})) formula
(\ref{eq:Eikonal_surf_difform}) can be interpreted as the inner product
of the action gradient $\mathbf{S}_{v}\left(v\right)\times\mathbf{P}\left(v\right)$
and the endpoint variation $\delta\mathbf{S}\left(v\right)$ both
are the elements of some function (-triplet) space:
\begin{gather}
\delta S=\left\langle \nabla S,\delta\mathbf{S}\right\rangle ,\:\nabla S\left(v\right)=\mathbf{S}_{v}\left(v\right)\times\mathbf{P}\left(v\right).\label{eq:Eikonal_surf}
\end{gather}
Summarizing the result of our examination on the minimal surface problem:
we need to find minimal paths in the space of boundary contours, \textit{i.e.}
in the space of coordinate function triplets -- or alternative contour
representation with the same information content (see \textbf{Fig.}\ref{fig:onepath}).
The gradient $\mathbf{S}_{v}\left(v\right)\times\mathbf{P}\left(v\right)$
designates the normal at the momentary contour (at a particular point
on the momentary action surface). Again, by analogy, if $\delta\mathbf{S}\left(v\right)$
is parallel with this direction, that is $\mathbf{S}_{v}\left(v\right)\times\mathbf{P}\left(v\right)\parallel\delta\mathbf{S}\left(v\right)$
(for any parameter value $v$ along the momentary contour) then the
endpoint variation formula (\ref{eq:Eikonal_surf_difform}) can be
written so that it contains the point-wise magnitudes of the gradient
and the endpoint variation functions:
\begin{equation}
\delta S=\underset{v}{\oint}dv\left|\mathbf{S}_{v}\times\mathbf{P}\right|\left|\delta\mathbf{S}\right|,\label{eq:Eikonal_surf_difform-1}
\end{equation}
the formula suitable for the action map construction.

Author's warning: the equations derived for minimal surfaces - despite
it provides insight to many aspects of the minimal surface problem
- require rigorous mathematical foundation of the Banach space of
endcontours including but not limited to: the coherent definitions
for the basis set, norm, level sets, the normal spaces of the level
sets etc. All of these required for the construction of the static
action map - \textit{i.e.} the distance function - in the space of
endcontours, where the shortest distance between two contours is identified
with the surface area of the minimal surface spanned between them.

\subsection*{A.4 The Isotropic case}

First we consider the relation between the Euclidean and Riemannian
elementary area that is $\left|\mathbf{N}\right|dudv$ $\longleftrightarrow$
$\sqrt{\det^{\ast}\left[\mathbf{G}_{T}\right]}\left|\mathbf{N}\right|dudv$,
where $\det^{\ast}\left[\mathbf{G}_{T}\right]$ is the determinant
of the two-dimensional minor of the space metric expressed in the
tangent plane in any \textit{normalized} local coordinate system $\mathbf{s}_{u}$,
$\mathbf{s}_{v}$ with the normalization being $\mathbf{s}_{u}\times\mathbf{s}_{v}=\mathbf{n}$,
$\left|\mathbf{n}\right|=1$. (Without normalization the determinant
expresses the space metric with the parameterization dependent surface
metric blended in: $\det\left[\mathbf{G}_{T}\right]=$$\det^{\ast}\left[\mathbf{G}_{T}\right]\det\left[\mathbf{S}_{ik}\right]$,
$i,k\in\left\{ u,v\right\} $, where $\det\left[\mathbf{S}_{ik}\right]=\left|\mathbf{N}\right|^{2}$,
similarly to the length expression (\ref{eq:Riemann_length})). 

If the space metric is isotropic, the matrix of $\mathbf{G}_{T}$
can be given in orthonormal \footnote{Local orthonormal basis is chosen for illustration. No more normalization
necessary than introduced - neither orthogonal, nor unit lenght basis
vectors - as all return the same determinant value.} basis as $\left[\begin{array}{cc}
\Phi\left(\mathbf{S}\right) & 0\\
0 & \Phi\left(\mathbf{S}\right)
\end{array}\right]$ then the Riemannian area expression becomes $\Phi\left(\mathbf{S}\right)\left|\mathbf{N}\right|dudv$.
We consider this form as the simplest version of the geometric minimal
surface functionals, still suitable for segmentation. The Lagrangian
of the isotropic, inhomogeneous space is
\begin{equation}
L=\Phi\left(\mathbf{S}\right)\left|\mathbf{N}\right|.
\end{equation}
This form is the direct equivalent to (\ref{eq:isotropic_inhomogeneous}).
Similarly to the minimal path problem, the associated abbreviated
functional can be obtained by rearrangement:
\begin{gather}
S=\iint_{A}\Phi\left(\mathbf{S}\right)\left|\mathbf{N}\right|dudv=\iint_{A}\Phi\left(\mathbf{S}\right)\mathbf{n}\cdot d\mathbf{A}\nonumber \\
\rightarrow\mathbf{P}=\Phi\left(\mathbf{S}\right)\mathbf{n}
\end{gather}
 or using the special parameterization introduced above (including
the $\mathbf{S}_{v}\times\mathbf{n}\parallel\mathbf{S}_{u}$ from
(\ref{eq:Eikonal_surf_difform-1}))
\begin{equation}
S=\stackrel[a]{u}{\int}\underset{v}{\oint}\Phi\left(\mathbf{S}\right)\left|\mathbf{S}_{v}\right|\left|\mathbf{S}_{u}\right|dvdu.\label{eq:functional_isotropic}
\end{equation}
Its particular Eikonal's differential form is:
\begin{equation}
\delta S=\underset{v}{\oint}dv\,\Phi\left(\mathbf{S}\right)\left|\mathbf{S}_{v}\right|\left|\delta\mathbf{S}\right|.\label{eq:Eikonal_isotrop_difform}
\end{equation}

\bibliographystyle{plain}
\bibliography{refs}

\end{document}